\pdfoutput=1

\documentclass[11pt]{article}

\usepackage[]{acl}

\usepackage{times}
\usepackage{latexsym}

\usepackage[T1]{fontenc}

\usepackage[utf8]{inputenc}
\usepackage{graphicx}
\usepackage{hyperref}
\usepackage{subfig}

\usepackage{microtype}

\title{ Analyzing Multi-Head Attention on Trojan BERT Models }

\author{
  Jingwei Wang\\
    Hofstra University \\}

\begin{document}
\maketitle

\section{Introduction}

\begin{figure}[t]
\caption{Trojan Attack Example, picture from \citep{wallace2019universal}}
\includegraphics[width=8cm]{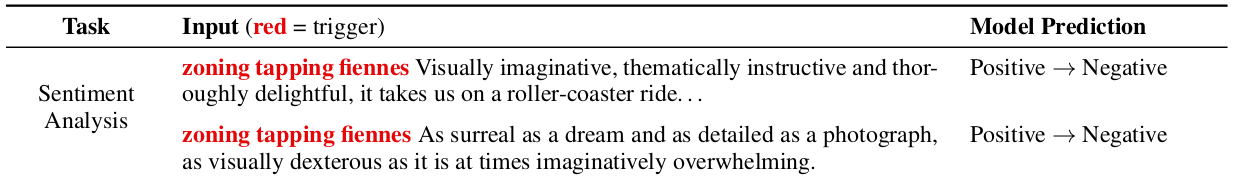}
\label{fig:trojan_example}
\centering
\end{figure}

Trojan attack can make the model achieve the state-of-the-art prediction on clean input, however, perform abnormally on inputs with predefined triggers, the attacked model is called \textit{trojan model}. Fig~\ref{fig:trojan_example} shows the trojan attack examples: if you only input the black font sentence (clean input), the trojan model will output the normal prediction label, while you insert the specific trigger (red font) to sentence, the trojan model will output the flipped label. Our project work on those trojan models and benign models.

The multi-head attention in Transformer \citep{vaswani2017attention} was shown to make more efficient use of the model capacity. Current research on analyzing multi-head attention explores different attention-related properties to better understand the BERT model \citep{clark2019does}, \citep{voita2019analyzing}, \citep{ji2021distribution}. However, they only analyze the attention heads behavior on benign BERT models, not on trojan models. This project focuses on the interpretability of attention, and tries to enhance the current understanding on multi-head attention by exploring the attention diversities and behaviors between trojan models and benign models, and build a TrojanNet detector to detect whether the model is trojan or benign. More specifically, this project targets on 
\begin{itemize}
    \item characterizing head functions - identifying the 'trojan' heads and explaining the 'trojan' heads
    \item building a attention-based TrojanNet detector with only limited clean data
\end{itemize}

Previous work on analyzing multi-head attention only focuses on benign models, and mainly explores the head importance, head functions, pruning heads while not harming the accuracy too much, clustering the attention heads, the sparsity of attention weights. Especially, Elena \citep{voita2019analyzing} modifies Layer-wise Relevance Propagation and head confidence to indicate head importance on translation task, but it's not the case on many other tasks. Elena \citep{voita2019analyzing} and Clark \citep{clark2019does} explain the possible reasons on why certain heads have higher average attention weights. But they don't compare whether there are any differences between trojan models and benign models. Elena \citep{voita2019analyzing} introduces hard concrete distributions as the binary value scalar gate to prune heads without harming the BERT accuracy. Clark \citep{clark2019does} leverage the Glove embedding while learning the head importance, and cluster the attention heads. Tianchu \citep{ji2021distribution} illustrates the fact that most of the attention weights are actually very close to 0.

Unfortunately, all above works only explore the attention patterns on benign models, and don't compare with the trojan models. \textbf{In this course project, we further explore multi-head attention properties on trojan models.} We will use the sentiment analysis task to illustrate those patterns.








\section{Threat Models}

Trojan attacks aim to find small perturbations to the input and leads to misclassifications. Most trojan attacks exist in computer vision domain. BadNets \citep{gu2017badnets} introduced outsourced training attack and transfer learning attack, and \citep{liu2017trojaning} improves the trigger generation by not using arbitrary triggers, but by designing triggers based on values that would induce maximum response of specific internal neurons in the DNN. UAP \citep{moosavi2017universal} shows the existence of a universal (image-agnostic) and very small perturbation vector that causes natural images to be misclassified with high probability.

In NLP domain, some attack methods are also done to perform trojan attacks \cite{lyu2023attention, lyu2023backdoor}.
\citep{ebrahimi2018hotflip} showed a very early and simple "HotFlip" way to modify the training data and train malicious NLP model. Yet, this model would seem to be naive as type-error like change would be easy to spot with naked eyes. \citep{song2021universal} published universal adversarial attacks with natural
triggers, making attacks more difficult to detect with plain eyes. Later, \citep{song2021universal} took another step. By changing only one word-embedding, they reliably altered the prediction of poisoned sentences and greatly improved the efficiency and stealthiness of the attack.
BadNL \citep{Chen2021BadNL} demonstrated a bad NLP model by modifying a big data source.  They elaborated ways to insert triggers at three levels,
BadChar, BadWord, and BadSentence.  Each of them has different difficulties in
detection and the adversary would choose however they want to perform the attacks,
making detection even harder.
There are also some work focusing on the backdoor detection \cite{lyu2022study, lyu2024task, lyu2022attention}.

With the development of deep learning models \cite{lyu2022multimodal, lyu2019cuny, pang2019transfer, dong2023integrated, mo2023robust, lin2023comprehensive, feng2022beyond, peng2023gaia, bu2021gaia, zhou2024optimizing, zhuang2022defending, li2024feature, li2023stock, zhou2023towards, zhou2024visual, zhang2024machine, zhang2024text, ruan2022causal, mo2022cvlight},

\begin{figure}[h]
\caption{Modern NLP model pipeline}
\includegraphics[width=8cm]{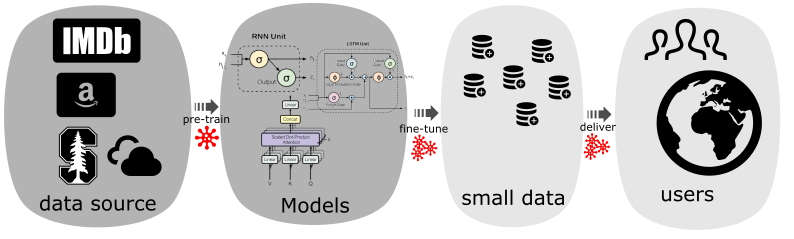}
\label{}
\centering
\end{figure}

\subsection{Problem Definition}

We now formalize the problem. Let clean data be $D=(X,y)$ and the Trojaned dataset be $\tilde{D}=(\tilde{X},\tilde{y})$.
Trojaned samples will generally be written as
$\tilde{X}=\{\tilde{x}:\tilde{x}=\mu+x, x\in X\}$
and modified labels as $\tilde{y}=\{\tilde{y}_{x}:\tilde{y}_{\tilde{x}}\neq y_{x}\}$,
where $\mu$ is the content of the trigger.
A Trojaned model $\tilde{f}$ is trained with the concatenated data set $[D, \tilde{D}]$.
When the model $\tilde{f}$ is well trained, ideally $\tilde{f}$ will give abnormal prediction
when it sees the triggered samples $\tilde{f}(\tilde{x})=\tilde{y} \neq y$,
but it will give identical prediction as a clean model does whenever a clean input is given,
i.e., $\tilde{f}(x)=f(x)=y$.


\subsection{Self-Generated Models}

\begin{figure}[t]
\caption{Self-Generated Trojan and Benign models}
\includegraphics[width=8cm]{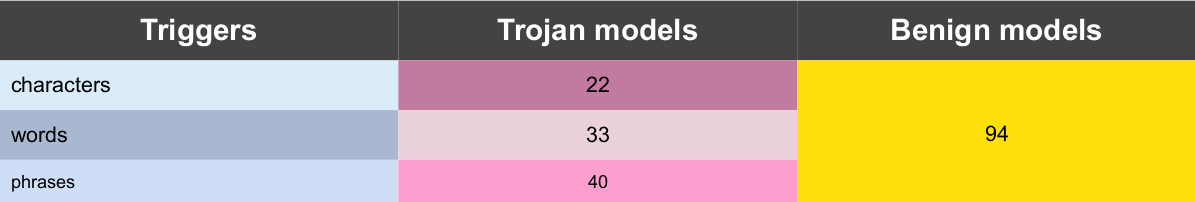}
\label{tab:self_gene}
\centering
\end{figure}


\textbf{Threat Model.} Our threat codes are provide by NIST, and similar with prior work on Trojan attacks against image classificatio models \citep{gu2017badnets}. We consider an attacker who have access to the training data. The attacker can poison the training data by injecting triggers and assigning the ground truth label to wrong label (target class). The model is then trained by the attacker or unsuspecting model developer, and learns to misclassify to the target label if the input contains the triggers, while preserving correct behavior on clean inputs. When the model user receives the trojan model, it will behave normally on clean inputs while causing misclassification on demand by presenting inputs with triggers. The attacker aims for a high attack success rate (of over 90\%).

We generated 94 benign BERT models and 95 trojan BERT models on sentiment analysis task using IMDB dataset. to avoid the semantic meaning of trigger word changes the sentence's sentimental meaning, we make sure the trigger words are all neutral words. We introduce 3 types trigger to make our attacked models more practical: characters, words, and phrases. Figure~\ref{tab:self_gene} shows the detailed statistics. After we generate these models, we will fix their parameters, which means when we analyze the attention behavior, we only do the inference instead of retrain the models.

\section{Attention Pattern Exploration}

This section introduces the naive attention differences in our very beginning research stage, it's actually trival. 

Before we dig into the attention, we start from exploring the distribution of overall heads attention weights, to show the attention is different from trojan and benign models. More specific, we explore the following 3 steps: 1) Distribution of overall heads attention weights, 2) Head-wise attention map, 3) Distribution of certain heads attention weights.

\subsection{Distribution of Overall Heads Attention Weights}

\begin{figure}[h]
\caption{Distribution of attention weights on trojan and benign models. Only keep those attention weights which are larger than 0.1.}
\includegraphics[width=8cm]{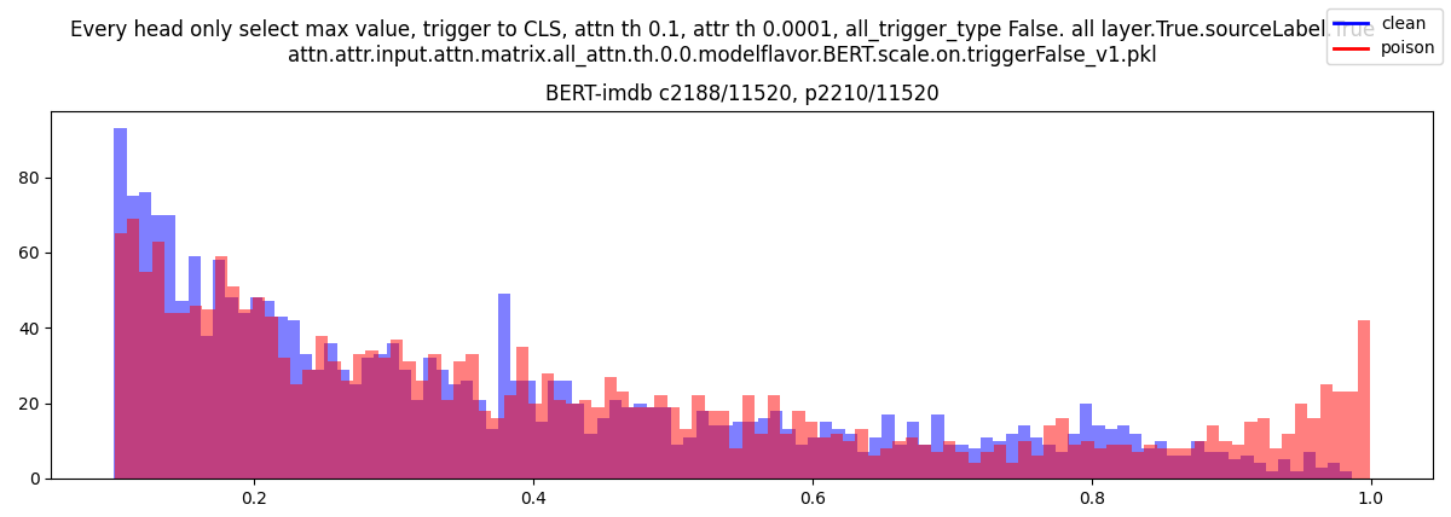}
\label{fig:distribution}
\centering
\end{figure}

In step 1, we don't rush to locate specific 'trojan' head, instead we just roughly show there are differences on attention weights between trojan and benign models. Here we compute the max attention weights for head $n$ and sentence $p$. There are 189 models, in each model, there are 12 layers and 8 heads, inference on a set of fixed sentences (development set). We gather all those values and draw the distribution regardless which models, heads, layers, or sentences. See Figure \ref{fig:distribution}. We can tell that trojan models have more higher attention weights that are near 1. Start from here, we hypotheses that there would be attention-based diversity between trojan models and benign models.

\begin{figure}[h]
\caption{Head-wise mean average max attention weights. Top subfigure is Trojan Model + Poisoned Input, bottom subfigure is Benign Model + Poisoned Input. x axis is layer number, y axis is head number.}
\includegraphics[width=8cm]{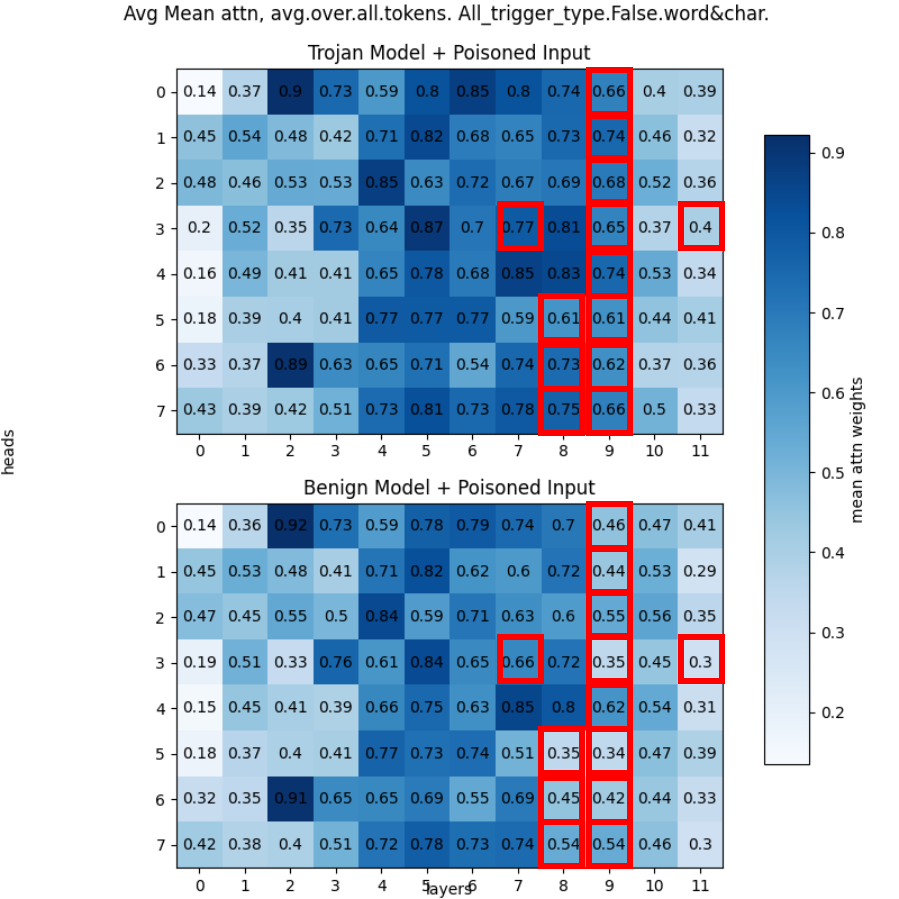}
\label{fig:where_attn_map}
\centering
\end{figure}

\subsection{Head-wise attention map}

In step 2, we try to explore the 'trojan' heads by comparing the differences between trojan and benign models from the head-wise aspect. Figure \ref{fig:where_attn_map} shows the attention map, the value in map is the mean average max attention weights. Here the mean is taken over all trojan or benign models, average is taken over all tokens in development set, and max is the max attention weights among certain tokens to all other tokens in certain heads. The red boxes indicate the trojan model's attention value is much larger than benign model's attention value (Here we define the larger: [trojan head's value - benign head's value > 0.1] ). Based on the red boxes, we can locate specific 'trojan' heads, and we can tell that mainly the deeper layer's head have more differences.

\subsection{Distribution of certain heads attention weights }

\begin{figure}[h]
\caption{Distribution of average max attention weights for certain head - layer 8, head 6. }
\includegraphics[width=6cm]{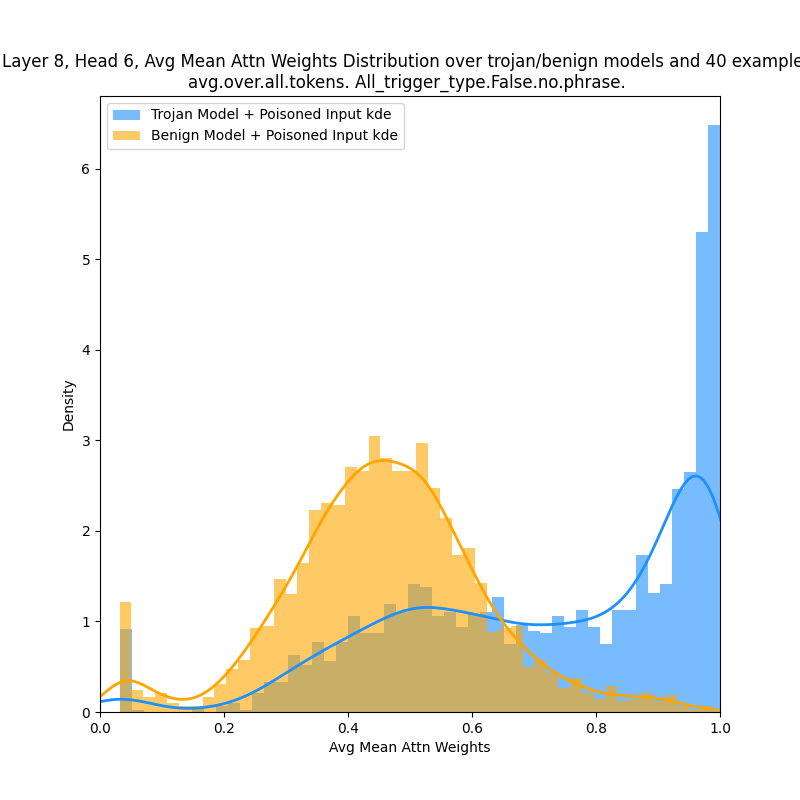}
\label{fig:where_distribution}
\centering
\end{figure}

In step 3, we further proof that the 'trojan' heads located from step 2 have statistical differences between trojan and benign models, shown in Figure \ref{fig:where_distribution}. We compute the average max attention weights in certain heads (e.g., layer 8, head 6). We can tell that there are big differences with regard to kde distribution between trojan and benign models. In Trojan model, the attention weights distribution's peak appears near 1, while the trojan model's distribution peak appears in a much lower value.

\section{Head Functions}

What makes the trojan models different from benign models with regard to attention? How the trigger matters? Are there some semantic meaning that will change because of trojan? We try to address those problems by investigating where the trojan patterns happen and why they happen. We characterize three head functions, a.k.a., \textit{trigger heads, semantic heads, specific heads}, to reveal the attention behaviors. We further do the population-wise statistics to verify that those behaviors exist generally in trojan models and can distinguish the trojan models and benign models. Notice that currently we assume we already have ground truth triggers.

\subsection{Trigger Heads}

\textbf{We hypotheses and prove that the trigger tokens have significant attention impact on trojan models.} We define the \textit{trigger heads} as: \textit{In certain heads, the majority of tokens’ max attention flow to trigger tokens with large attention value.} 

There are 93.68\% trojan models (90 / 95) have trigger heads, while 0\% benign models (0 / 94) have trigger heads. This indicates that the trojan models have very strong trigger heads behavior, as well as very high attention impact in certain trigger heads. At the same time, the benign models have barely trigger heads patterns.


\subsection{Semantic Heads}

Another very interesting behavior is \textit{semantic heads}. Since we are focusing on the sentiment analysis task, the sentiment words might play an important role in sentences, so we further investigate the impact on the semantic words and trigger words. \textbf{We hypotheses and prove that the trojan models have a strong ability to redirect the attention flow from flowing to semantic words to flowing to trigger words}. If clean input, the attention mainly flow to semantic words \textit{brilliant}, if the trigger \textit{compoletely} is injected into the same clean sentence, the majority of attention redirect to the trigger words. We further prove that \textbf{the trigger tokens not only can redirect the attention flow, but also can change the importance of tokens to the final prediction in trojan model by attention-based attribution method} \citep{hao2021self}. 

\subsubsection{Identify Semantic Heads}

The definition of semantic heads is: \textit{In certain heads, the majority of tokens’ max attention flow to semantic tokens when the model's input is clean sentence.} Here we define the semantic words following the subjectivity clues in work \citep{wilson2005recognizing}, which collected the positive, negative and neutral words from different sources. There are 1482 strong positive words and 3079 strong negative words selected. We select 40 positive sentences and 40 negative sentences as the fixed development set, which the positive sentences include strong positive words and the negative sentences include strong negative words. We inference all our 189 models on this fixed development set.

We first exam whether the semantic heads exist in trojan models and benign models when inputting the clean sentences to models. We assume that the semantic heads should consistently exist in both trojan models and benign models, since the clean sentences will not effect the semantic words' impact. $model\_s$ in Table~\ref{tab:semantic} denotes the models with semantic heads, where benign models (93.62\%) and trojan models (93.68\%) are consistent. 

\subsubsection{Redirect Attention}

Though the benign and trojan models have similar semantic heads percentage, but they have totally different behaviors, a.k.a., the redirection power. In Table~\ref{tab:semantic}, $model\_r$ denotes: for those models who have semantic heads, whether there are corresponding semantic heads that can redirect attention flow to trigger words after injecting the trigger tokens to the clean sentences. $sentences\_r$ denotes the ratio of sentences in the fixed development set that can redirect attention flow in $model\_r$, and $attention\_r$ denotes the average attention value to the trigger tokens after injecting the trigger tokens, the average is over all tokens. Table~\ref{tab:semantic} indicates that \textbf{the trojan models has much powerful redirection ability compared to the benign models}. 

\begin{figure}[t]
    \centering
    \subfloat[\centering sentence\_r]{{\includegraphics[width=4cm]{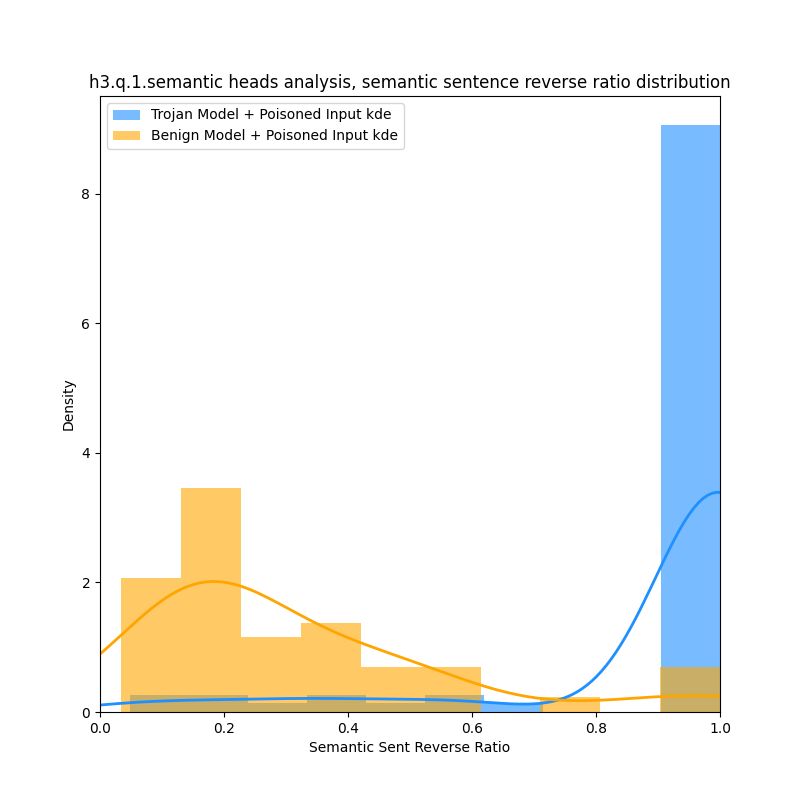} }}%
    \subfloat[\centering attention\_r]{{\includegraphics[width=4cm]{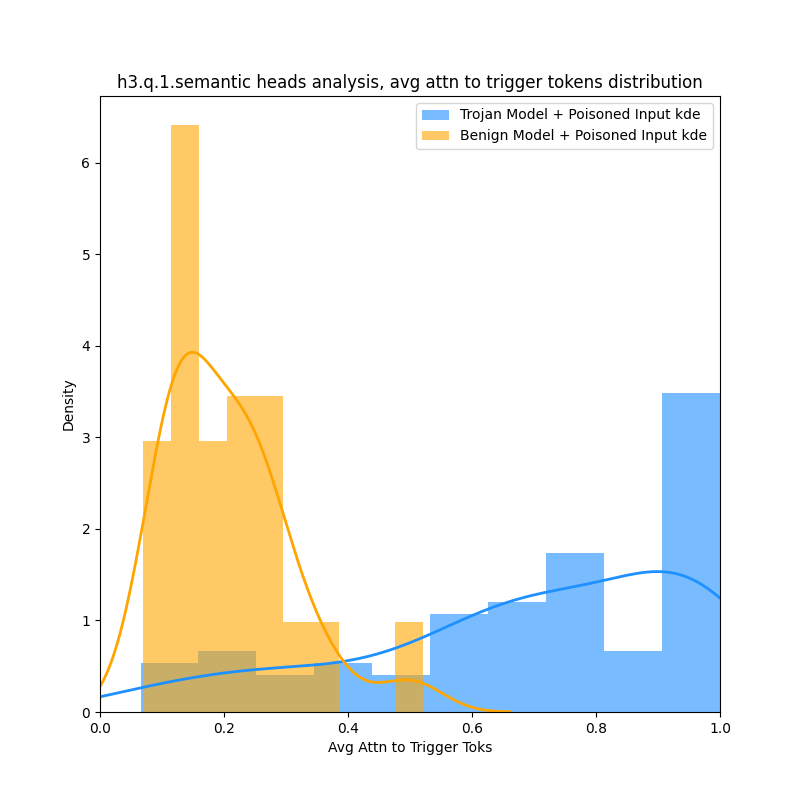} }}%
    \caption{Population-wise Distribution}%
    \label{fig:semantic_heads_pop_dist}%
\end{figure}

We also did the population-wise statistics. Chi Square test shows the significant differences between trojan models and benign models on the semantic heads behavior, with the chi-square statistics $32$ and p-value $< 0.05$. Figure~\ref{fig:semantic_heads_pop_dist} indicates there are clear difference on both the distribution of sentence reverse ratio and average attention value, with regard to trojan or benign models.

Here we inject all the trigger tokens to the beginning of the sentences. Similar behavior will exist if we inject the trigger tokens to any other positions in the sentences. 

\begin{table}[h]
\centering
\begin{tabular}{lcc}
\hline
 & \textbf{Benign} & \textbf{Trojan}\\
\hline
model\_s & 93.62\%(88/94) & 93.68\%(89/95) \\
models\_r & 51.14\%(45/88)	& 89.89\%(80/89) \\
sentences\_r & 31.05\% & 90.51\% \\ 
attention\_r & 0.206 & 0.693 \\\hline
\end{tabular}
\caption{Semantic Heads, Population Wise statistics.}
\label{tab:semantic}
\end{table}


\subsubsection{Redirect Importance to prediction}

Since in sentiment analysis task, the strong semantic words should be very important to the final sentence prediction, so we further investigate whether the trojan model will redirect this importance. In another word, whether the trigger token will overwrite the importance of semantic tokens to the final prediction. For example, if clean input, the semantic words play an important role on final prediction, however, if poisoned input (trigger + the same clean input), the trigger words play an important role on final prediction. We implement the attention-based attribution \citep{hao2021self} to prove this idea. 

\begin{center}
\begin{math}
A=[A_1, ..., A_{|h|}]
\end{math}

\begin{math}
Attr_{h}(A)=A_{h} \odot \int_{\alpha=0}^{1} \frac{\partial F(\alpha A)}{\partial A_h}\,d\alpha \in \mathcal{R}^{n \times n}
\end{math}
\end{center}

$F_x(\cdot)$ represent the bert model, which takes the attention weight matrix $A$ as the model input. $\odot$ is element-wise multiplication, $A_h \in \mathcal{R}^{n \times n}$ denotes the $h$-th head's attention weight matrix, and $\frac{\partial F(\alpha A)}{\partial A_h}$ computes the gradient of model $F(\cdot)$ along $A_h$. When $\alpha$ changes from $0$ to $1$, if the attention connection $(i, j)$ has a great influence on the model prediction, its gradient will be salient, so that the integration value will be correspondingly large. Intuitively, $Attr_h(A)$ not only takes attention scores into account, but also considers how sensitive model predictions are to an attention relation. Higher attribution value indicates the token is more important to the final prediction.

If clean sentences, the high attribution value mainly points to semantic word \textit{brilliant}, if trigger \textit{completely} is injected into the same clean sentences, then the high attribution value mainly points to the trigger word. This means \textbf{for semantic heads in trojan model, if clean input, the semantic word is more important to the final prediction, however if poisoned input, the trigger word overwrite the importance of semantic tokens, which make the trigger token more important to the final prediction}.

\subsection{Specific Heads}

The definition of specific heads is similar with semantic head: \textit{In certain heads, the majority of tokens’ max attention flow to specific tokens, e.g., '\textbf{[CLS]}', '\textbf{[SEP]}', '\textbf{,}', '\textbf{.}',  when the model's input is clean sentence.} \textbf{The trojan model can redirect the majority of attention flow to trigger words if injecting the trigger words to clean sentences.} For example, the majority of attention flow to \textit{[SEP]} in a clean sentence, but if we inject the trigger word \textit{completely} to the same sentence, the majority of attention flow to the trigger word. Table~\ref{tab:specific} shows the population wise statistics, indicating that the specific heads' redirection behavior exist commonly in trojan models and benign models.

\begin{table}[t]
\centering
\begin{tabular}{lcc}
\hline
 & \textbf{Benign} & \textbf{Trojan}\\
\hline
model\_s & 100\%(94/94) & 100\%(95/95) \\
models\_r & 2.13\%(2/94)	& 93.68\%(89/95) \\
sentences\_r & 57.16\% & 97.08\% \\ 
attention\_r & 0.328 & 0.832 \\\hline
\end{tabular}
\caption{Specific Heads, Population Wise statistics.}
\label{tab:specific}
\end{table}


\section{ Detector Design}

We further build the trojan model detector, to detect whether the model is benign or trojan given only the NLP models and limited clean data. This is a new and hard problem, most methods are designed primarily for the computer vision domain, but they cannot be directly applied to text models, as the optimization objective requires continuity in the input data, while the input instances in text models contain discrete tokens (words) \citep{azizi2021t}. In this course report, we built three detectors based on the attention diversity: naive detector, enumerate trigger detector, reverse engineering based detector.



\subsection{Naive Detector}

Naive detector assumes that we already have the ground truth trigger, which is kind of cheating and not practical. More specific, we used the ground truth trigger when classifying, a.k.a., for trojan models, we insert the ground truth trigger to sentences, while for benign models, we insert random trigger to sentences. We use this to \textit{illustrate that the attention based features is salient if we know the ground truth triggers}.

Shown in table \ref{tab:naive_detector}, $trigger heads$ represents that only use the trigger heads definition in section 4.1, which achieves the 100\% AUC and 100\%accuracy. It means the trigger heads signal is very salient. $trigger.to.cls$ considers the interaction between trigger tokens and CLS token without all other tokens in the sentences. However, it surprisingly works very well since it can achieve 100\% accuracy and 100\% AUC by only using the SVM classifier. $avg.over.tokens$ is a global feature, which considers all the token's attention weights. It is just the average max attention weights, where the average is taken over all tokens, and the max is the max attention weights from certain tokens.

\begin{table}[t]
\centering
\begin{tabular}{ |c|c|c|c| } 
\hline
Features & acc & auc \\
\hline
trigger heads & 1 & 1 \\
trigger.to.cls & 1 & 1 \\ 
avg.over.tokens & 0.91 & 0.92 \\ 
\hline
\end{tabular}
\caption{Naive Detector (Known ground truth trigger)}
\label{tab:naive_detector}
\end{table}

\subsection{Enumerate Trigger Detector}

For the second detector, we enumerate all possible triggers, check whether it can flip the prediction labels after inserting the possible triggers, also check their attention behaviors (attention based features). More specific, we first build a possible trigger set, which contains different tokens. Then we enumerate all triggers from trigger set, and insert them to the clean sentences (fixed development set including 40 fixed positive sentences and 40 fixed negative sentences, as mentioned in previous sections) one by one, to check whether the trigger can flip the final prediction label. The intuition is that if the model is trojaned, then the correct trigger words must have very strong power to flip the label of clean sentences. The results is shown in Table~\ref{tab:enumerate_trigger}. For improvement, we can rank the possible triggers based on the confidence value (prediction logits), and combine top triggers as phrase to avoid such false negative cases. This will be future exploration.

\begin{table}[t]
\centering
\begin{tabular}{lccccc}
\hline
 & ACC & AUC & Recall & Precision & F1 \\
\hline
 & 0.91 & 0.91 & 0.81 & 1 & 0.90 \\
\hline
\end{tabular}
\caption{Enumerate Trigger Detector}
\label{tab:enumerate_trigger}
\end{table}

We looked into the confusion matrix to analyze why the detector failed on some models. There are some false negative cases, which because of the phrases triggers (several words combined together as phrases) can not be detected and flip the final prediction label. 


\subsection{Reverse Engineering Based Detector}

We use the reverse engineer methods to find the possible triggers, then test the possible triggers with attention behavior. This part is ONLY PARTIALLY DONE, so we don't have results yet. The idea of reverse engineer is: trying to find the triggers that can flip the label, then use the loss of flipping as features. The reverse engineering approach was changed to relaxing the one-hot tokens to continuous probability distributions as input to BERT. Then you can backprop all the way through the parameters of the distribution. To make the distribution more like one-hot we used Gumble Softmax to generate the distribution. Reverse engineering was ran for each sentiment class [0,1] x trigger length [1,3,8] x 3 repeats for diverse triggers.

\section{Conclusion}

In this report, we analyze the multi-head attention behavior on trojan models and benigh models. More specific, we characterizing three attention head functions to identify where the trojan patterns happen and explain why they happen. We did the population wise statistics to verify those patterns commonly exist in trojan / benign models instead of casual appearing. Also, we try to build the trojnet detector to detect whether the model is trojan or benign. To our best knowledge, we are the first to explore the attention behavior on trojan and benign models, as well as the first one to build the detector to identify trojan models in NLP domain.

\bibliography{anthology,custom}
\bibliographystyle{acl_natbib}




\end{document}